\documentclass[atmp]{ipart_v1}

\Vol{NN}
\Issue{1}
\Year{2025}
\firstpage{701}

\usepackage{t1enc}
\usepackage[latin1]{inputenc}
\usepackage{soul}
\usepackage[english]{babel}

\usepackage{amsthm}
\usepackage{yfonts}

\usepackage{bbm}
\usepackage{bm}
\usepackage{mathrsfs}
\usepackage{xcolor}
\usepackage{graphicx}

\newcommand{\BZ}{\mathbb{Z}}
\newcommand{\BR}{\mathbb{R}}

\newcommand{\Tr}{\mbox{Tr}}
\newcommand{\Id}{{\bf 1}} 

\newcommand{\be}[0]{\begin{equation}}
\newcommand{\ee}[0]{\end{equation}}
\newcommand{\bea}[0]{\begin{eqnarray}}
\newcommand{\eea}[0]{\end{eqnarray}}
\newcommand{\E}[1]{\mathbb{E}\left[#1\right]}
\newcommand{\EE}[2]{\mathbb{E}_{#1}\left[#2\right]}

\newcommand{\CO}[0]{{\mathcal O}}
\newcommand{\CA}[0]{{\mathcal A}}
\newcommand{\CN}[0]{{\mathcal N}}

\numberwithin{equation}{section}

\theoremstyle{plain}

\def\boldclass{\bf\sf}

\def\NP{{\boldclass NP}}

\begin{document}

\title[Diffusion Models for Cayley Graphs]{Diffusion Models for Cayley Graphs}

\author[Michael R. Douglas and Kit Fraser-Taliente]{Michael R. Douglas and Kit Fraser-Taliente}


\begin{abstract}
We review the problem of finding paths in Cayley graphs of groups and group actions, using the Rubik's cube as an example,
and we list several more examples of significant mathematical interest.
We then show how to formulate these problems in the framework of diffusion models.
The exploration of the graph is carried out by the forward process, while finding the target nodes is done by
the inverse backward process.  This systematizes the discussion and suggests many generalizations.  To
improve exploration, we propose a ``reversed score'' ansatz which  substantially improves over previous 
comparable algorithms.

\end{abstract}

\maketitle

\section{Introduction}

Machine learning systems are increasingly able to learn to solve difficult problems through exploration, without the need for an explicit algorithm or even a full description of the problem. 
A classic example is to solve Rubik's Cube: starting from an arbitrary position, find a sequence of
moves which brings the cube back to a simple starting position. 
There are many algorithmic solvers such as Cube Explorer,\footnote{
{\tt https://kociemba.org/cube.htm}. }   
whose designers encoded domain-specific knowledge about the Rubik's cube. 
Can a computer learn to solve
the cube without being given such knowledge, much as AlphaZero learned to play go and chess?  
This challenge was first met in \cite{mcaleer2018solving,agostinelli2019solving}.

Among mathematical puzzles, Rubik's cube is particularly admired as a clear illustration of the concepts of
group theory.\footnote{A reader unfamiliar with the cube should look at \cite{bandelow2012inside}.}
There are 12 elementary Rubik's cube moves -- choose a face (6 choices), and do a quarter twist of the 9 cubelet
block containing that face.  Each move has an inverse (the opposite quarter twist), and composition of moves obeys
the associative law.  Thus, any composition of moves defines a group element, and the complete set defines the Rubik's Cube
group $G_{Rubik}$, a finite group with order (number of elements)
$2^{27} 3^{14} 5^3 7^2 11 \sim 4\times 10^{19}$.  Despite this large order, any position can
be returned to the starting position in at most 26 moves.\footnote{
The oft-quoted ``God's number'' of 20 counts half-twists as single moves.
}

One can define $G_{Rubik}$ as follows.
Give distinct labels to each of the $6\times 8$ facets (leaving
out the center facets which don't move); then each position is related to the starting position by a permutation of these labels;
thus an element of the permutation group $S_{48}$.  $G_{Rubik}$
is the subgroup of $S_{48}$ obtained by taking all products
of any number of the 12 elementary moves.\footnote{A more explicit description is given in \S \ref{s:graphs}.}

The construction we just described is an example of defining a group $G$ in terms of a subset $S\subset G$ of generators. 
If every $g\in G$ can be obtained as a finite product of elements from $S$, we say $G$ is generated by $S$.
The choice of $S\subset G$ is additional structure (compared
to just specifying $G$), and making it allows us to
define the Cayley graph $\Gamma(G,S)$.
This is a graph whose vertices are group elements $g\in G$ and whose edges connect pairs of vertices $(g,g')\in G\times G$ such that $ga=g'$ for some $a\in S$.
Thus one can phrase the Rubik's cube problem mathematically as, given a starting point $g\in G$, find a path in $\Gamma(G,S)$ to the identity element $\Id$.  
A candidate algorithm for solving the problem is given by specifying
a ``policy'' function $\pi:G\rightarrow S$, 
the next move to take if one is in the position $G$.
Letting $g_{t+1}=g_t \pi(g_t)$ defines a path starting at $g_{t=0}$,
and the algorithm is a solution if every path reaches $g_T=\Id$
for some $T$.\footnote{  
One might ask for more: to find a path between any two elements $g,g'\in S$,
but this can be solved by
finding a path from $g^{-1}g'$ to $\Id$.}

\subsection{Generalized Rubik's cubes}

This statement of the problem does not refer to any particulars of the original
Rubik's cube, only $\Gamma(G,S)$, so we can regard it as defining
an infinite set of ``generalized Rubik's cube'' or group navigation problems.
Here are a few of the Cayley graphs for which mathematicians have studied this: 
\begin{itemize}
    \item $G=SL_2(\BZ)$, the $2\times 2$ matrices with integer entries and unit determinant.
    We take the set of generators to be $S=\{T_{\pm 1},U_{\pm 1}\}$ with
\be\nonumber
    T_{k} = \left(\begin{matrix}1 & k \\ 0 & 1\end{matrix}\right); \qquad
    U_{k} = \left(\begin{matrix}1 & 0 \\ k & 1\end{matrix}\right)
\ee
    \item  $G=SL_2(\BZ_p)$, the group of $n\times n$ invertible matrices over the finite field $\BZ_p$, 
    and $S=\{T_{\pm 1},U_{\pm 1}\}$.
    \item Still taking $G=SL_2(\BZ_p)$ but with $S$ a slightly larger generating set,
    one gets the LPS Ramanujan graphs \cite{lubotzky1988ramanujan}, the first constructions of expander graphs. 
\end{itemize}
One could also take $G=SL_q$, or many other variations.  Many examples from combinatorics are in 
\cite{konstantinova2008some}.
In physics, the problem of designing a quantum circuit can be put in this form, taking $G$ to be a finite subgroup of $U(N)$ and $S$ to be the operations realized by quantum gates \cite{Lin_2019}.  Later we will generalize to Cayley graphs of group actions, to include many more problems
 such as the Andrews-Curtis problem.

Algorithms for these problems are known and they range widely in difficulty.  Before discussing this, we should be more precise in our statement of the problems, as so far we did not
say how elements of $G$ are presented as inputs to the policy $\pi$.
For these matrix groups, it is natural to take the matrix elements (as $q^2$ elements of $\BZ$ or $\BZ_p$) as the input.  Then, the first problem ($G=SL_2(\BZ)$) is easy to solve using the
Euclidean algorithm for gcd (see appendix \ref{app:euc}).
The second is difficult and the third is even \NP\ complete
(considered as a family of problems labeled by $p$ and $q$)  \cite{t2021complexity}.

This representation was a choice and in general one could use as input to the algorithm any function $F:G\rightarrow \BR^n$
into some space of features.   For the Rubik's cube, one usually uses a permutation representation;
the $(i,j)$ feature is $1$ if the $i$'th label is in the $j$'th position and $0$ otherwise.
There are other generalizations of the problem,
for example to finding paths in a graph derived from
the action of a group acting on a set.  We discuss these in section \S \ref{s:graphs}, as well as the question:
can every graph be put in this form (answer: no), and what is special about Cayley graphs which 
 makes navigation problems more solvable and/or more learnable.

\subsection{Reinforcement learning solvers}

Let us now discuss solvers which are not given any information about the group or graph to start with,
rather they must learn about it by exploration.  This puts us squarely in the realm of reinforcement learning
(RL).  The state space is $G$, the set of actions is $S$ and there is a reward for reaching
the goal $\Id$.  One can then apply one of the many RL methods.  DeepCube \cite{mcaleer2018solving}
trained a value-policy network, while DeepCubeA \cite{agostinelli2019solving} trained a network
to learn an estimate of the ``cost to go'' function, the cost to reach the goal.

Here, this cost function is simply the length of the shortest path in $\Gamma(G,S)$ from a specified element $g$ to the identity.  Call this length $\ell^*:G\rightarrow\BZ$, and call the estimate $\ell:G\rightarrow\BZ$.
We can use this to formulate a ``greedy'' policy: 
if we are at $g$, we simply compute $\ell(g)$ and $\ell(ga)$ for $a\in S$, and choose the action $a=\pi(g)$ which minimizes $\ell(ga)-\ell(g)$.\footnote{This is not always the best approach.
Consider the $SL(2,\BZ)$ problem above: the algorithm in appendix \ref{app:euc} is more easily implemented
by predicting the policy $\pi$.}
One can also use more sophisticated search algorithms, as we discuss below.

The exploration and learning of $\ell(g)$ can be made systematic by generating random walks starting
from the identity.  Thus, we generate a forward walk $g_0=\Id$ and $g_t=g_{t-1} a_t$, where the steps
$a_t$ are sampled uniformly from $S$.  One can then minimize an error $|\ell(g_t)-\ell(g_{t-1})-1|$.
This is only approximate as it assumes that each step increases the distance (no backtracking) but
works well for the Rubik's cube.  

A simpler prescription \cite{chervov_machine_2025}
is to learn the functional relation $\ell(g_t)=t$.  This is also approximate if considered
as a means of learning $\ell^*$, but it is exact (by definition) if we reinterpret it as learning the expected
time for a random walk starting at $\Id$ to reach $g$.  Now, the greedy policy works just as well if
$\ell(g)$ is a monotonic function of $\ell^*(g)$, which is plausible for 
the expected time to reach $g$.

Another (earlier) variation on this \cite{takano2023self} is to 
train a predictor for the last move $a_t$ which was applied to reach $g_t$.
The most likely last move is generally the move most likely to decrease the distance to the goal,
so this also gives us a policy.  The predictor estimates the probability distribution
$p_t(g_t,a_t)$ for the last move in a random walk reaching $g_t$, by minimizing the cross
entropy between it and the distribution of sampled random walks.  The solver then
organizes the search through candidate paths from a starting point $g$ by weighing them according
to the predicted probability that each choice will move towards the goal.

\subsection{Our contributions and outline}

This brings us to the new idea of our work.  
We propose to unify the data generation process and the search process as the forwards and
backwards random walks of a diffusion model.  Diffusion models \cite{sohl2015deep} are a very successful
paradigm for generative AI, which originated in concepts from physics.  This new point of
view will simplify the discussion, will make contact with relevant results from mathematics and
physics, and will suggest new generalizations to the algorithm.
For example, there is no {\it a priori} reason why the data generating process (the forward
walks) should sample the action space uniformly.  One could use a general diffusion process,
and vary it to optimize the exploration.  We will propose a
concrete version of this idea, which performs better in experiments.


In \S \ref{s:graphs}, we review a few basic definitions: graph theory, Cayley graphs, graphs of group
actions, and related concepts.  While our proposal makes sense for general graphs, we explain the 
additional structure present in these more restricted classes of graphs.  We also explain some
outstanding mathematical problems which can be formulated as finding paths in Cayley graphs of group actions and/or
showing that the graphs are disconnected.

In \S \ref{s:diff} we explain the relation between graph navigation and diffusion, and make
our proposal.  The exploration and search functions are implemented in terms of forward and backward 
random walks.  Following the diffusion model approach, we take these to sample from inverse diffusive processes,
and give an objective function for learning the backward process.  Using these backward walks, one can search by
sampling independent backward walks, or one can use these in more sophisticated algorithms.
We discuss beam search from this point of view, and comment on other algorithms.

In \S \ref{s:endtoend} we discuss modifying exploration by modifying the forward process.  We define
expected search time and use it to define the quality of the system.  We discuss end-to-end
training schemes based on this as an objective function.  We then propose a ``reversed score'' ansatz relating the forward 
and backward processes which is simple to implement, and argue that this will also promote exploration.

In \S \ref{s:experiments} we discuss experimental results for the Rubik's cube and for a $SL_2(\BZ_p)$ problem
using the reversed score proposal, which performs as well or better than previous systems.
We conclude in \S \ref{s:conclusions}.

\section{Concepts and definitions}
\label{s:graphs}

In the following we adopt notation that, although applicable in a broader context, is particularly suited to Cayley graphs.
We consider a directed graph $\Gamma(G,S^G)$, where $G$ is the set of vertices.
Here $S^G$ denotes a function which, given a $g\in G$, produces the set $S^g$ of edges $a\in S^g$ which start at $g$.
We denote the vertex reached by following the edge $a$ leaving $g$ as $g\;a$, and a path starting at $g$ as
$g\; a_1 \; a_{2}\; \ldots \; a_k$.
We require each edge to have an inverse, so given $g\in G$ and $a\in S^g$ there exists a unique
$a^{-1}\in S^{ga}$ with endpoint $g$.\footnote{
One could redo the discussion in terms of undirected graphs if desired.
Also, the subsequent proposals can be straightforwardly generalized to multigraphs, weighted graphs and the like.
However general directed graphs are not allowed, as in this case
the inverse for the forward diffusion does not take the form Eq. (\ref{eq:backwards}).
}

Let the distance $d(g,h)$ be the length of the shortest path from $g$ to $h$, so $\ell(g)=d(g,\Id)$.
The diameter of $\Gamma$ is the maximum distance for any pair $(g,h)$.
The adjacency matrix $A$ of the graph is a symmetric $|G|\times |G|$ matrix with
entries $A_{g',g}=1$ if $\exists a\in S^g$ s.t. $g'=ga$ and $0$ otherwise.
The degree matrix $D$ is a diagonal $|G|\times |G|$ matrix with $D_{g,g}=\mbox{deg}\, g = |S^g|$,
the number of edges leaving $g$.

The Cayley graphs $\Gamma(G,S)$ defined in the introduction are examples.  Here $S\subseteq G$ is a fixed subset
so we can simplify the notation.   They are highly symmetric examples: 
intuitively, the graph is ``the same'' around every vertex.
More precisely, for every $g,h\in G$ there is a graph automorphism $L$ such that $L(g)=h$.
These automorphisms define an action of the group $G$ on the graph (or $G$-action, about which more below).
This action is transitive (every pair $g,h$ is so related)
and free (the only solution to $L(g)=g$ is the identity automorphism).
These are defining properties for Cayley graphs of groups: any graph with a group $G$ of automorphisms with 
these two properties is a $\Gamma(G,S)$ for some $S$.

Of particular relevance for
our discussion, the problem of finding a path from any $g$ to any goal $h$ can be reduced to that for the goal $\Id$.\footnote{
A path $g a_1 a_2 \ldots a_k=h$ is also a path $h^{-1} g a_1 a_2 \ldots a_k=\Id$.
}
Thus $d(g,h)=d(h^{-1}\; g,\Id)$.  We can relax some of these constraints to obtain larger classes of graphs with
some but not all of this structure.

One such class are the Cayley graphs of more general group actions.
A right action of a group $G$ on a set $X$ is a group homomorphism $G\rightarrow \mbox{Aut}(X)$, or
more concretely a map $A:G\times X\rightarrow X$ such  that
$A_{g'}(A_g(x))=A_{gg'}(x)$ for all $g,g'\in G$ and $x\in X$.
Given $A$ and a set of generators $S\subset G$, we define the Cayley graph of the action $A$,
denoted $\Gamma_A(X,S)$
(or just $\Gamma(X,S)$ if this is clear), as a directed graph with vertices $X$ and edges
labeled by pairs $(x,a)\in X\times S$, which take $x$ to $x\;a\equiv A_a(x)$.

In these terms, the original Cayley graph $\Gamma(G,S)$ with edges from $x$ to $x\; a$ is the case in which 
$A_a(x)=R_a(x)=x\;a$, an action of $G$ on itself called the right regular action.
One can also define the left regular action $L_g(x)=g^{-1}\;x$.\footnote{
The inverse is there to match with our definition of right group action above.}
This action gives rise to automorphisms of $\Gamma(G,S)$, the group $G$ of automorphisms discussed above.

Cayley graphs of group actions are far less constrained.  
This is the case even if $|X|=|G|$.  Different points $x\in X$ need not be related by symmetries,
so the navigation problem depends on both starting and ending points. 
The action need not be transitive, so $\Gamma(X,S)$ can be disconnected.
It might have fixed points $x$ ($A_g(x)=x \,\forall g$) or stabilizers
(the same but only for $g\in H\subset G$).  Deciding whether these possibilities
are realized is often interesting.

A few examples of group actions and associated mathematical problems:
\begin{itemize}
\item As an elementary example, consider the action of spatial rotations on a cube.  The set $X$ could consist
of assignments of labels to the vertices, edges and/or faces.
If we restrict to symmetries of the cube, a rotation will act by permuting these labels.
While the choices involved give different representations of the group, if the action is free and
transitive, these are all equivalent to the Cayley graph problem for the symmetry group.

\item Let's come back to the standard Rubik's cube and think more carefully about the labels \cite{chen2004group}.
There are 8 corner cubelets which can be permuted among themselves.  Each cubelet has 3 labels attached in a fixed cyclic
order, which can be rotated into 3 configurations.  
Similarly there are 12 permutable cubelets shared by pairs of faces,
with 2 labels on each.  Let $X$ be the set of these choices, so $|X|=8! \times 3^8\times 12! \times 2^{12}$.
The Rubik's cube twists $a\in S$ act on $X$, so there is a graph $\Gamma(X,S)$.
However this is {\bf not} the same as $\Gamma(G,S)$, because not all arrangements of the
cubelets can be produced by twists of the starting position (some are ``invalid'').  
In particular,
the operation of pulling out a single cubelet, turning it to rotate its labels
and replacing it in the cube cannot be undone by a twist of the cube.  Cube twists also
act the same way on the parity of the two permutations (in $S_8$ and $S_{12}$).
Thus the $G$-action is not transitive, and $\Gamma(X,S)$ has 12
connected components ($G$-orbits).  Each orbit is isomorphic to $\Gamma(G,S)$,
so $|G|=|X|/12$ (reproducing the order of $|G|$ quoted above).  

\noindent
Our RL solver represents cube positions as permutations of labels; thus one can give
it an invalid position and ask it to find a path to the starting position.  But there 
is no such path.  What will it  do?  Will it find a path to a nearby position and stop?
Does the trained model give us any clue that $\Gamma(X,S)$ is disconnected? 

\item Consider the $SL_2(\BZ)$ problem above but now with $S_k=\{T_{\pm k},U_{\pm k}\}$
for $k>1$.  Such an $S_k$ only generates a subgroup $H\subseteq SL_2(\BZ)$, but this subgroup
still acts by right multiplication on all of $X=SL_2(\BZ)$.  Thus there is a
graph $\Gamma(X,S_k)$, and one can ask the analogous questions, in particular whether
this graph is connected.

\item The Markoff triple problem 
(as explained to us by Jordan Ellenberg \cite{ellenbergblog}).  
We take $G=SL_2(\BZ)$ and $S=\{T_{\pm 1},U_{\pm 1}\}$.
But instead of acting on vectors in $\BZ^2$, we define a $G$-action $\CA$ on the free group with two generators $\{A,B\}$ as follows.
Given a pair of words $u$ and $v$ defined as products of $A^{\pm 1}$ and $B^{\pm 1}$, let
$T_{\pm 1} : (u,v) \rightarrow (uv^{\pm 1},v)$ and $U_{\pm 1} : (u,v) \rightarrow (u,u^{\pm 1}v)$.
This is a $G$-action on $F_2$, the free group with two generators
(words in $\{A^{\pm 1},B^{\pm 1}\}$).  
These actions also preserve the commutator $uvu^{-1}v^{-1}$, for example $T_{+1}$ takes this to $(uv) v (uv)^{-1} v^{-1}$.

Now, given any homomorphism into another group
(say an explicit choice of group elements representing $A$ and $B$) we get
a $G$-action on its image.
For this problem, we are interested in representations of $\{A,B\}$ as matrices in a second $SL_2(\BZ)$, say 
$A = \left(\begin{matrix}1 & 1 \\ 1 & 2\end{matrix}\right)$,  
$B = \left(\begin{matrix}2 & 1 \\ 1 & 1\end{matrix}\right)$.
These matrices satisfy $\Tr ABA^{-1}B^{-1}=-2$ and therefore (one can show)
the commutator is unipotent ($(ABA^{-1}B^{-1}+\Id)^2=0$).  
And since the action $\CA$ preserves the commutator, every 
pair of matrices $A',B'$ obtained by acting on $A,B$ by the $G$ action $\CA$
will satisfy $\Tr A'B'A'^{-1}B'^{-1}=-2$.

Does the converse hold: can every $A',B'$ with this property (trace $-2$ commutator)
be obtained from $A,B$ by the action $\CA$?
Taking $X$ to be the set of matrices with this property, this is the
same as asking: is $\Gamma_\CA(X,S)$ connected?
This turns out to be true \cite{bourgain_markoff_2015}, and these
authors asked: is this also the case if we consider the analogous action 
on $SL_2(\BZ_p)$ ?
In \cite{Chen2020NonabelianLS,Eddy2023ConnectivityOM} this was shown to be true for all but
finitely many primes $p$.
In a sense, the mod $p$ problems
determine the solvability of the integer problem, a phenomenon of great interest in number theory.  This problem is also a warmup for:

\item The Andrews-Curtis problem.
We follow the problem statement given in  \cite{borovik_finitary_2011,romankov_andrews-curtis_2023}.
Consider a group $G$, which can be general but
which we will eventually take to be $F_k$ the free group
with $k$ generators.  Let $G^k$ be the set of $k$-tuples of elements of $G$.
Define the normal closure $NC_G(R)$ of a subset $R\subset G$ to be the group generated
by all $grg^{-1}, g\in G, r\in R$.  And define $N_k(G) \subset G^k$ to be the
normal generating sets with $k$ elements, sets $R\in G^k$ such that $NC_G(R)=G$.

Now, there is a group of transformations $GAC_k(G)$, the general AC-group of $G$
of dimension $k$, 
which acts on subsets $R\in G^k$ in a
way that manifestly preserves $NC_G(R)$.  For example, the two sets
$(g_1,g_2,\ldots,g_k)$ and $(g_1,g_1^{-1} g_2,\ldots,g_k)$ generate the same group,
and the transformation that takes the first to the second is an element of $GAC_k(G)$.
If $G$ is finitely  generated by a set $S$, then $GAC_k(G)$ has a finite set
of generators $SAC_k(G,S)$, given on the first page of  \cite{borovik_finitary_2011}.

Since the group $GAC_k(G)$ preserves $NC_G(R)$, it acts on $N_k(G)$.  Given a set of generators 
$SAC_k(G,S)$ this action has a Cayley graph $\Gamma_{GAC_k(G)}(N_k(G),SAC_k(G,S))$.
It is called the AC-graph $\Delta^S_k(G)$.

The Andrews-Curtis conjecture is that $\Delta^S_k(F_k)$ is connected
(one can take $S$ to be any set of generators).\footnote{The original statement of the problem
was that every presentation of the trivial group can be brought to a canonical form.
The relation to the statement here is that $R\in N_r(G)$ can be interpreted as the relations
of a group presentation, and the quotient $F_k/NC_{F_k}(S)$ is trivial iff $S\in N_k(F_k)$.}
This was studied using theorem provers in \cite{davenport_andrews-curtis_2018}
and using RL in \cite{shehper_what_2024}.  In \cite{borovik_finitary_2011} 
(corollary 1.3) it was
shown that finite group analogs of this conjecture are true. 

We were able to use the techniques of \S \ref{s:experiments} to trivialize some simple group presentations, such as AK(2), the first element of the Akbulut-Kurby series of potential counterexamples.

\item Take the group $\mathbb{F}^{n^3}$, which is isomorphic to the additive group of $n \times n\times n$ tensors with entries in a number field $\mathbb{F}$. Define the generating set
$S = \{a \otimes b \otimes c \;|\; a, b, c \in \mathbb{F}^n, a \otimes b \otimes c\neq 0\}$, $|S| = 3|\mathbb{F}|^n$. Then the problem of identifying the rank of a tensor is equivalent to finding a shortest path on this graph. As an illustration of the utility of this problem, the tensor rank of the $9\times 9 \times 9$ tensor corresponding to the multiplication of two $3\times 3$ matrices is known \cite{ballard2018geometryrankdecompositionsmatrix} only to be in the interval $[19,23]$. A constructive demonstration of a new upper bound would give the fastest known matrix multiplication algorithm. 

Using the methods of \S \ref{s:experiments}, we were able to reproduce Strassen's solution of the $2\times 2$ problem for $\mathbb{F}=\mathbb{Z}_2$, by finding a length 7 path 
from the matrix multiplication tensor to the origin.

\item More general equational simplification or rewriting systems.  While one can formulate
simplification as a search problem on a graph, in general it is not a
    Cayley graph problem,    because rewriting operations usually do not have well defined inverses.

\end{itemize}

Not all pathfinding can be reformulated in terms of group actions.  
One requirement is that each action has an exact inverse, which requires it to be a bijection on $X$.
While there is a more general theory of semigroups (not requiring inverses), in this case the forward diffusion
operator might not have an inverse.  If it does, it need not take the form Eq. (\ref{eq:backwards}) with support on edges of the graph.
Another requirement is that the same set of actions can be applied to any state, and that there is
a global way to label the actions.  There is a more general theory of groupoids which does
not make these assumptions, for which the proposal does apply, but this theory is far less constraining.\footnote{
To illustrate the issues, consider a maze:  states $X$ are a subset of $\BZ^2$, say, a finite region with
points deleted, and the actions are the 4 unit moves, call these $(L,R,U,D)$.  There is a graph $\Gamma(X,S^X)$ and
a navigation problem, but since unit moves into deleted points are not allowed, $S^X$ is not the same for every state $x\in X$.
Nor can it can be turned into a group action by redefining the action of such illegal moves to keep the location fixed.
The problem with this definition is that several locations map
to the same new location and thus this action has no inverse.
One gets a semigroup, see \cite{liu_transformers_2022} for examples.}

\section{Diffusion model}
\label{s:diff}

Consider a graph $\Gamma(X,S)$ with a distinguished (or ``target'') element $\Id\in X$.
The navigation problem is, given a vertex $x\in X$, find a path to $\Id$.
A generalization is to give a set of target elements $E\subset X$, and a solution
is a path to any $e\in E$.

We will generate training data using random walks starting at $\Id$, but now we think of this as sampling from
a discrete time diffusive Markov process. Such a process has a stationary distribution which is uniform over the graph. This suggests thinking of the reverse path as the inverse of this diffusion, concentrating the stationary distribution onto the goal state.

Given a position on a graph $x_t \in X$, for $t \in \mathbb{Z}, 0\leq t\leq T$, one defines transition probabilities $q_{t+1|t}(x_{t+1}|x_t)$. Taking $T$ steps yields a trajectory $X_{0:T}$. Suppose that we sample $x_0$ from some initializing distribution $x_0 \sim p_0$, the set of 'goal states'. Denoting the probability of such a trajectory as $p_{0:T}(x_{0:T})$, write:

\begin{equation} \label{eq:forwards}
p_{0:T}(x_{0:T}) = p_0(x_0)\prod_{t=1}^{T} q_{t|t-1}(x_{t}|x_{t-1})
\end{equation}

This is a discrete time\footnote{We can also formalize this process in continuous time.} Markov process which we call the forward process. In practice, $q$ has very simple structure. For the case of interest, we give it by the appropriately normalized adjacency matrix $D^{-1}A$ of $G$, and it is time-homogeneous. We can write down the reverse Markov process via an application of Bayes' rule:

\begin{align} \label{eq:backwards}
p_{0:T}(x_{0:T}) &= p_T(x_T)\prod_{t={T}}^{1} \tilde{q}_{t-1|t}(x_{t-1},x_{t}),\\  \label{eq:backwardskernel}
\tilde{q}_{t-1|t}(x_{t-1},x_{t})  &= \frac{q_{t|t-1}(x_{t}|x_{t-1}) p_{t-1} (x_{t-1})}{p_{t}(x_{t})}
\end{align}

This defines a time-inhomogeneous Markov process with the transition kernel $\tilde{q}_{t-1|t}$. If we knew $\tilde{q}$, one could first sample from $p_T$, and then draw samples from $p_0$ by applying the transition $\tilde{q}$ at each time step. For the case of a Cayley graph with generators $S$, we only need to establish $\tilde{q}$ on $[0,T]\times G\times S$ because $\tilde{q}$ is proportional to $q$.

It is therefore sufficient to learn $\frac{p_{t-1} (x_{t-1})}{p_{t}(x_{t})}$, which in the literature is referred to as the \emph{score}\footnote{Consider the traditional notion of the score $\nabla_x\log p(x)$, but replace the regular derivative with the discrete derivative on graphs.}. If we learned the score, we would be able to straightforwardly reverse the diffusive process. We propose using a neural network to approximate:

\begin{equation} \label{eq:score}
\sigma_{\theta,t}(x_{t-1},x_t) \approx  \frac{ p_{t-1} (x_{t-1})}{p_{t}(x_{t})}
\end{equation}

In training such a neural network, a natural choice of objective function would be the expectation value of the MSE between the two. Clearly, this is not the correct choice for positive-definite probability-like distributions, and moreover the distributions are not normalised. We proceed by using an objective function \cite{lou2024discrete} based on a Bregman divergence $D_F(\cdot,\cdot)$. The Bregman divergence is nonnegative, symmetric, and convex divergence determined by a strictly convex function $F$. We use\footnote{Here $y|x$ denotes that $y\neq x$ and are connected by a single edge.}

\begin{equation} \label{eq:defL}
\mathcal{L}[\sigma_\theta]=\sum_t\mathbb{E}_{x\sim p_t}\sum_{y|x} D_F\left(\sigma_{\theta,t}(y,x),\frac{p_{t-1}(y)}{p_t(x)}\right)\frac{p_{t-1}(y)}{p_t(x)}, \quad F = -\log
\end{equation}

\begin{align}
D_F(f,g) &= F(f)-F(g) - F'(g)(f-g),\quad F = -\log\\
D_{-\log}(f,g) \cdot g &= f - g\log f + (g\log g - g)
\end{align}

After subtracting a $\theta$-independent constant,\footnote{The constant is simply $\mathbb{E}_{y\sim p_{t-1}}\log{\frac{p_{t-1}(y)}{p_t(x)}}$. In the case when $\Gamma$ is a Cayley graph, it can be interpreted as the sum over the KL divergences between $p_{t-1}(gx)$ and $p_t(x)$ for all $g$. It is positive-definite and bounded strictly away from zero.} the `loss' can be rewritten in a way which is easily calculable.
\begin{equation} \label{eq:loss}
\mathcal{\mathcal{L}'}[\sigma_{\theta}] = \sum_{t=1}^T \left(\mathbb{E}_{x\sim p_t} \sum_{y| x} \sigma_{\theta,t}(y,x) - \mathbb{E}_{x\sim p_{t-1}}\sum_{y|x} \log \sigma_{\theta,t}(x,y)\right).
\end{equation}


While we made the discussion having a Cayley graph in mind, so far everything we said makes sense for an arbitrary undirected graph $\Gamma$.\footnote{
If we are willing to use a score function of two arguments $x,y$ which is not always well defined.
We suspect this might be problematic for learning and generalization.
}
If we now assume the set of actions $S^x$ is independent of $x\in X$, 
we can take the score function Eq. (\ref{eq:score}) to be a function $\sigma_{\theta,t}(x)_a = \sigma_{\theta,t}(y=xa, x)$,
and the loss becomes
\begin{equation}\label{eq:cgloss}
\mathcal{\mathcal{L}'}[\sigma_{\theta}] = \sum_{t=1}^T \left(\mathbb{E}_{x\sim p_t} \sum_{a} \sigma_{\theta,t}(x)_a - \mathbb{E}_{x\sim p_{t-1}}\sum_{a} \log \sigma_{\theta,t}(xa)_{a^{-1}}\right).
\end{equation}
At the cost of additional variance, one can replace the sum over $t$ with $\mathbb{E}_t$.


We give diffusion Algorithm 1 for graph exploration and navigation.
\begin{itemize}
    \item To train the model, generate random walks of length $T_f$ starting at $\Id$ (or any target $e\in E$)
    using the forward process Eq. (\ref{eq:forwards}),
    and learn a score function by gradient descent on Eq. (\ref{eq:loss}) or Eq. (\ref{eq:cgloss}).
    \item Given a position $x\in X$, to find a path back to $\Id$ (or $E$)
    generate a random walk according to the backwards process Eq. (\ref{eq:backwards}) starting at time $T_b=T_f$
    and decreasing the time: so $x_T=x, x_{T-1}=x a_{T-1}$, {\it etc.}.
    If the walk reaches $E$, stop, otherwise continue until $t=0$. 
\end{itemize}

The choice of $T_f$ is constrained by the need to explore the entire graph (so $T_f > \mbox{diameter}$),
and on the other side by the nature of diffusion: as $T$ increases, eventually the probability distribution
and the score will converge on the uniform distribution, so the signal will be washed out. If $T$ is chosen much too large, therefore, one spends a significant amount of time learning--for the most part--the uniform distribution.

If one were to learn the true score $\sigma^*$, following the reverse diffusive process would guarantee reaching the goal state $x_0$ at time $0$, assuming the initial state is indeed within a ball of radius $T$. Moreover all \emph{full} trajectories $x_{[0:T]}$ - that is, paths \emph{conditioned} on starting and ending at $x_0$ at time 0 and $x_T$ at time $T$ - are equally probable. This is true whether one is considering the forward or reverse process, by definition. Running the reverse process starting from $x_T$, in general one should expect to select a random length-T trajectory between $x_T$ and $0$.

There are many variations on the algorithm.  One evident possibility is to, instead of implementing the reverse search as a random walk, instead take the most probable action at each time step. Alternatively, the search algorithm could use multiple samples from the backwards distribution: one sensible strategy for doing this is beam search: an algorithm which keeps $N$ random walks $x_t^i$, all starting at $x_{T_b}=x$:
\begin{itemize}
    \item For each walk $x_t^i$, sample a backwards step $\alpha$ times, to get $\alpha N$ walks.
    \item Use Eq. (\ref{eq:backwards}) to compute the probability of each of these walks, and keep the
    $N$ most probable.
\end{itemize}
We will see empirically that this improves the results (finds shorter paths), particularly when paired with an optimal solve for a small ball of radius $R$ around the origin. 
But before presenting these results, we introduce a further variation on the algorithm.

\section{Improving the forward process}
\label{s:endtoend}

Now that we have reformulated the data generation process as a diffusion process, it is clear that generating the data by
making uniform random walks was a choice.  We could have chosen a different transition matrix $q_{t|t-1}(x_{t}|x_{t-1})$
from the start, and we could also consider varying the forward process during the exploration to more efficiently explore the graph.

It is easy to see that the uniform random walk is not always the most efficient data generating process.
Consider $G$ abelian, then backtracking walks (those containing both $a$ and its inverse $a^{-1}$) add no additional value.
On the other hand if we know nothing about $G$ to start, then there is no grounds on which to start with a non-uniform walk;
any improvements would have to be learned.  Thus we can pose the problem of learning a modified forward diffusion $q^*$ defined by
\begin{equation} \label{eq:qstar}
p_{0:T}(x_{0:T}) = p_0(x_0)\prod_{t=1}^{T} q^*_{t|t-1}(x_{t}|x_{t-1}) ,
\end{equation}
by simultaneously training it with the score on a loss which combines Eq. (\ref{eq:loss})
with an additional term which measures the quality of the solution.
Let us first define such a term.

Given a search process governed by a policy $\pi$,
we define the expected time $\hat T_b$ to reach the goal as a function
of $x\in X$ as
\be \label{eq:exptimex}
\hat T_b(x) = \E{ t+T_{penalty}\, \bigg| x_{\pi,T-t}(x) = \Id}
\ee 
where $x_{\pi,t}(x)$ is the random path starting at $x_T=x$ generated by $\pi$.
Normally $T_{penalty}=0$, but if the time reaches $t=T$ without reaching $x=\Id$, 
the walk stops and the penalty is applied.
Given a distribution $\mu_{start}(g)$ of starting points, the expected solution time is then   
\be \label{eq:exptime}
\hat T_b(\mu) = \int \mu_{start}(g) \hat T_b(g) .
\ee
For a finite graph it is natural to take the uniform distribution, but this is a choice.
Either $\mu_{start}$ is given to the algorithm in some way, or it needs to be learned.

We now have two objective functions.
Eq. (\ref{eq:defL}) depends implicitly on the forward process $q^*$ (through sampling) and also on $\sigma$, and enforces the
inverse relation between the processes. The $\sigma$ dependence is explicit, so it is straightforward to do gradient descent on $\sigma$. We can also gradient-descend on $q^*$ by writing down an estimator for the gradient with respect to the parameters of $q_\theta$, precisely as in the usual `policy gradient' trick for reinforcement learning.
Varying $q^*$ changes the `$\theta$-independent' term we dropped from Eq. (\ref{eq:defL})
to get Eq. (\ref{eq:loss}) $\left(\sum_g D_{KL}(p_{t-1}(x g)/p_t(x)\right)$, where $p$ depends on the forward process). One way to deal with this issue 
is to alternately train $\sigma$ and $q^*$.  We could also decide to simply ignore this term (perhaps because $q^*$ does not vary that much), and only training $\sigma$
is also worth trying\footnote{One could imagine using a PPO-type KL divergence limitation whilst training $q^*$ to ensure that it doesn't change too quickly, although this is probably not be necessary.}

Eq. (\ref{eq:exptime}) depends on the backward process and thus on $\sigma$ and $q^*$. 
Now neither dependence is explicit, but we can estimate the dependence by using the multiple paths
of the beam search.  Suppose we have a set $W$ of walks with varying $T_b$, then we compute the probability of the
$i$'th walk as a functional of $\sigma$, and estimate $\hat T_b$ as a functional of $\sigma$ as the expected $T_b$ over $W$.
In more detail, we have a set $S$ of $N$ backwards random walks $\{x_t^i\}$ starting at $x_T$ of which some have reached a goal.
We want to use them to estimate $\hat T_b(\sigma)$.  If they had all reached the goal, this would be straightforward, we
would write
\be\label{eq:esthatTb}
\hat T_b \sim \EE{S}{t|x_{T-t} \in E}
\ee
where 
\be
\EE{S}{A} \equiv \CN \sum_{x^i \in S} P[x^i]\; A[x^i]
\ee
where the probability $P[x^i]$ is given by Eq. (\ref{eq:backwards})
and the normalization $\CN$ is chosen so that $\EE{S}{1}=1$.
This can be varied with respect to the parameters of the backward process $q^*$ and/or $\sigma$
to get a gradient.

We need to decide how to treat backward walks which did not reach the goal. One possibility is to leave them out.
Another option would be to take the final point $x^i_t$ and use $\hat T_f(x^i_t)$, the expected time to
reach $x^i_t$ with the forward walk, as an estimate of the remaining $\hat T_b(x^i_t)$.  Finally, we
could define such walks as not having reached the goal and count them as $\hat T_b=T_{penalty}$.  This
is correct at $t=T$ and simple to implement.

This leads to the following Algorithm 2:
\begin{itemize}
    \item Generate random walks of length $T_f$ starting at $\Id$ (or any target $e\in E$)
    using the forward process Eq. (\ref{eq:qstar}),
    and improve the score function $\sigma$ by gradient descent on Eq. (\ref{eq:loss}) or Eq. (\ref{eq:cgloss}).
    \item Generate backward walks starting at $x_T$ obtained by sampling from $\mu_{start}(x)$ using
    Eq. (\ref{eq:backwards}) as in Algorithm 1.   Measure $\hat T_b$ and improve $q^*$ to decrease $\hat T_b$. 
\end{itemize}

Finally, one could consider adding other terms to the objective function.
For example, one could add a measure of the expected variance of the random walk, as the best walks for both
exploration and search are the direct walks to the target or goal.
The entropy of $\sigma_t(x)_a$ as a function of $x$ and $t$ might work.

It would be interesting to evaluate these proposals.  Our hunch is that Eq. (\ref{eq:exptime})
is too noisy a signal to give much improvement by itself, but the general idea seems promising.

\subsection{Reversed score ansatz}
\label{ss:revscore}

A simpler procedure is to define $q^*$ directly in terms of the score, which contains a lot of
information about the exploration process.  We propose an ansatz which defines the forward probability to traverse an edge $(x,a)$
as the score for the reverse edge $(xa,a^{-1})$, normalized to a probability.  Explicitly,
\begin{equation}\label{eq:ansatz}
q^*_{t+1|t}(x_{t+1} = xa| x_t = x) = \frac{\sigma^*_{t+1}(xa)_{a^{-1}}}{\sum_g\sigma_{t+1}^*(xg)_{g^{-1}}} =\frac{\frac{p_{t}(x)}{p_{t+1}(xa)}}{\sum_g\frac{p_{t}(x)}{p_{t+1}(xg)}}=\frac{\frac{1}{p_{t+1}(xa)}}{\sum_g\frac{1}{p_{t+1}(xg)}}.
\end{equation}
As seen in the final expression, this ansatz upweights exploration of less well sampled directions. 

In writing Eq. (\ref{eq:ansatz}) we used the true score $\sigma^*$.
Since this is determined by $q^*$ so that the forward and backward processes are inverses, and 
since the transition kernel at time $t$ is a function of the probability distribution at time $t+1$, 
this is an implicit (or self-consistent) equation which {\it a priori} may or may not have solutions. 
On a finite graph, there is at least one solution: the uniform distribution and uniform transition functions solve it with constant $p$ for all $t$ and $x$.

In practice, of course, we do not have access to $\sigma_t^*$, so we use $\sigma_{t,\theta}$, the neural network estimate of the score,
\begin{equation}\label{eq:qansatz}
q^*_{t+1|t}(x_{t+1} = xa| x_t = x) = \frac{\sigma_{t+1,\theta}(xa)_{a^{-1}}}{\sum_g\sigma_{t+1,\theta}(xg)_{g^{-1}}}
\end{equation}
and treat it iteratively.  Presumably, this iteration will converge to solutions of Eq. (\ref{eq:ansatz}), such as the uniform distribution.
Since it favors directions for which $p_{t+1}(gx)$ is small,
it seems plausible that it does so \emph{faster} than the uniform random walk.  Further analysis of this process is an interesting question for the future. 

This gives us training Algorithm 3, which iterates the following steps:
\begin{itemize}
    \item Generate random walks of length $T_f$ starting at $\Id$ 
    using the forward process Eq. (\ref{eq:qansatz}) determined by the current $\sigma$.
    \item Improve the score function $\sigma$ by gradient descent on Eq. (\ref{eq:loss}).
\end{itemize}

\section{Experiments}
\label{s:experiments}

\subsection{Results on the Cube}

We train a neural network with 16 million parameters for 100 million trajectories, using the improved forward process $q_\theta$ from Algorithm 3. The neural network is has four residual blocks, and uses LayerNorm. We used batches of 100 full trajectories $x_{0:T}$ (so effective batch length $100*T$). We took $T$ to be 30. We initialize trajectories from a ball of radius $N$ from the goal state, with the distance from the goal state uniformly selected from $[0,N]$ for each trajectory. We took $N$ to be 1. We trained, and ran inference, on a single NVIDIA RTX 4090 GPU. 

\begin{figure}
    \centering
    \includegraphics[width=0.5\linewidth]{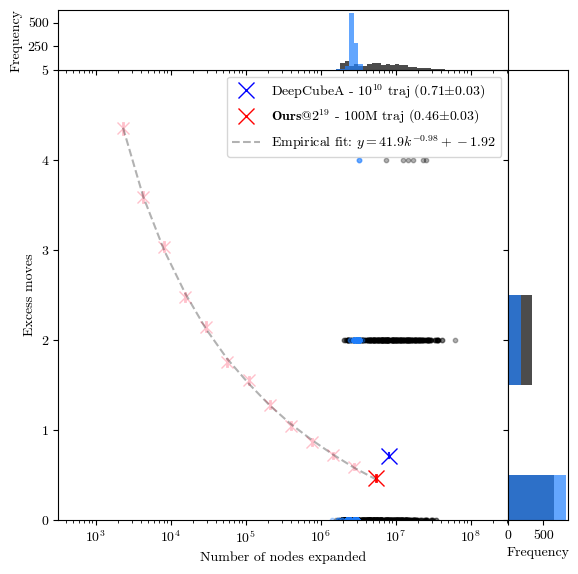}
    \caption{Average solution length vs beam width. We find a best fit given by $E = 41.9 log_2(B)^{-1}-1.92$, where $B$ is the beam width (trivially related to the number of nodes). Excesses are always even due to parity. We report results with a ball radius $R=5$, and no $T$-calibration.}
        \label{fig:beam}
\end{figure}

For beam search, we simply used the uniform forward process to determine the reverse kernel $\tilde{q}$ due to memory limitations. Empirically, this does not impact the performance of the beam search too much - it is more important to use it during training. Using an almost exactly comparable neural network to \cite{takano2023self,agostinelli2019solving} (modified only to add LayerNorm and introduct a sinusoidal time embedding), we achieve superior performance with many fewer examples seen: see Table \ref{tab:rubik}. We also compare to an optimal solver, which leverages domain knowledge and detailed knowledge of the Rubik's cube group to optimally solve states. It relies on 182 GB of memory, but is able to guarantee an optimal solution. We used this data to compute excess trajectory lengths, which are given in Fig. \ref{fig:beam}. We observed a smooth dependence, which was well fit by a power law $E \sim (\log_2 B)^{-0.98}$. Further analyzing this result may be a fruitful avenue for future work.

\begin{table}\label{tab:rubik}
\begin{tabular}{l|rr|rc|r}
\hline Method & Sol. length & Opt. (\%) & \# of nodes & Time taken (s) & T \\
\hline Optimal & 20.64 & 100.0 & ${2.05} \times 10^6$ & 2.20 &  \\
\hline Ours @ $2^{18}$ & $\mathbf{21.15(21.11)}$ & ${(76.6)}$ & ${2.76} \times 10^{6}$ & $10.09$ & 100M \\
\hline Ours @ $2^{19}$ & $\mathbf{21.06 (21.00)}$ & $\mathbf{(81.7)}$ & ${5.35}  \times 10^{6}$ & $19.93$ & 100M \\
\hline EC @ $2^{18}$& $21.26$ & ${69.6}$ & ${4.18}\times 10^{6}$ & $\emph{15.29}$ & 2B \\
DCA & $21.35$ & ${65.0}$ & $8.19 \times 10^6$ & \emph{28.97}& 10 B \\
\hline
\end{tabular}
\vspace{5mm}
\caption{We give our results for beam widths $2^{18}$ and $2^{19}$. In brackets, we report the improved score from \emph{$T$-calibration}, which doubles execution time but modestly improves the score with no additional training or memory requirements. Times in italics are estimated from the number of nodes expanded, as the times reported in the relevant papers are run on older hardware, presumably with a less efficient beam search implementation. EC is EfficientCube \cite{takano2023self}, DCA is DeepCubeA \cite{mcaleer2018solving}. DCA trained on 10B scrambled states, but did not train on complete trajectories, so this number may be misleading.}
\end{table}

\begin{figure}
    \centering
    \includegraphics[width=0.5\linewidth]{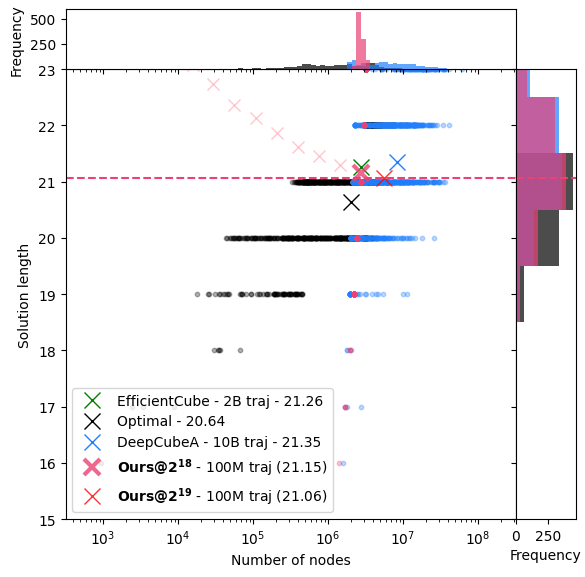}
    \caption{Average solution length against number of nodes expanded, for 1000 initial states randomly scrambled between 1000-10000 times. We report results with a ball radius $R=5$, and no $T$-calibration. For beam width $2^7$ or greater, we are able to solve all states.}
    \label{fig:sol}
\end{figure}

\begin{figure}
    \centering
    \includegraphics[width=0.9\linewidth]{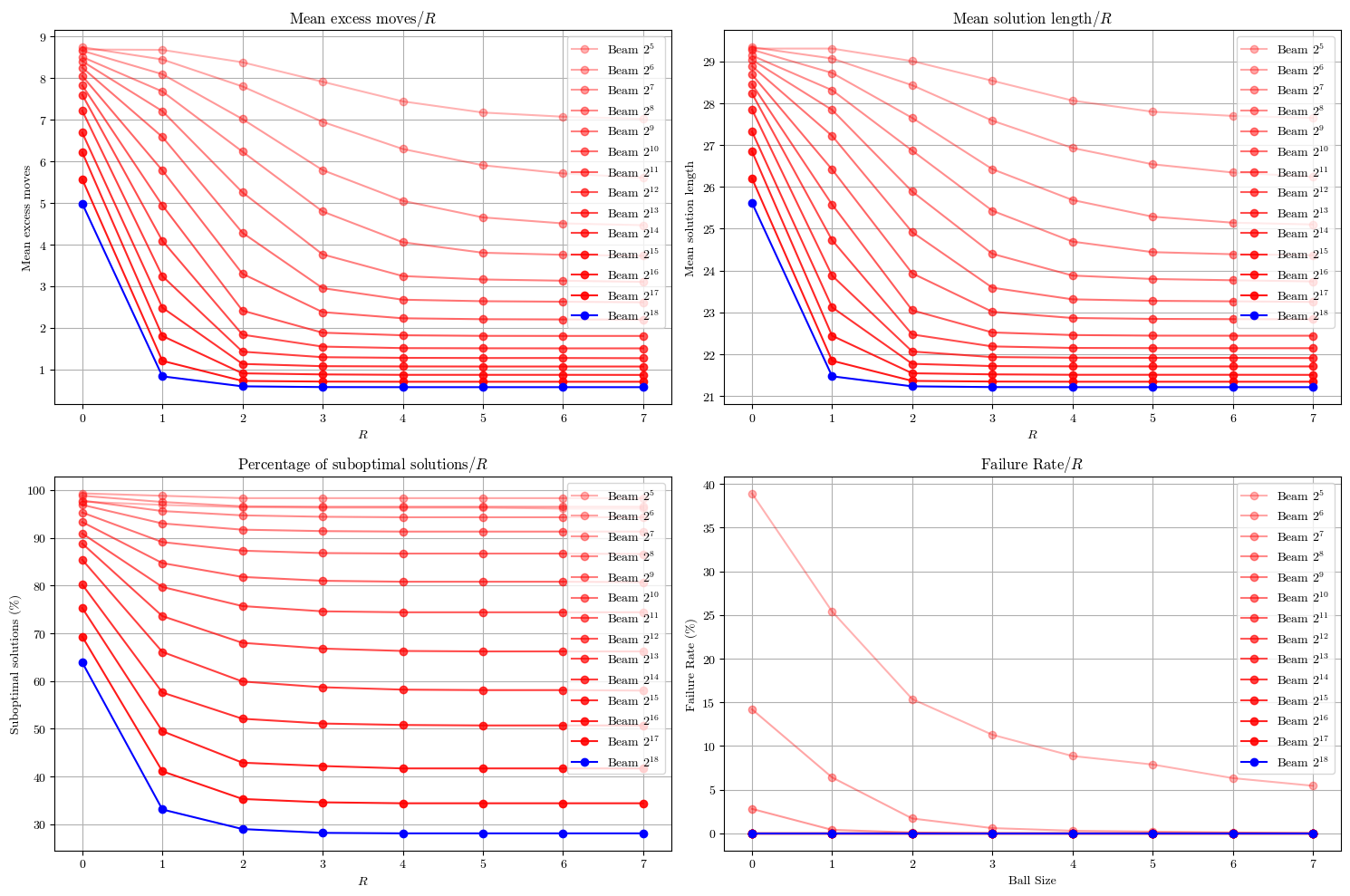}
    \caption{We observe smooth dependence of performance on beam width and radius of the hashed ball in the two-sided search. Of particular note is the significant degradation for zero ball size. Again, we conjecture this is related to the issue of $T$-calibration.}
    \label{fig:scoresvsball}
\end{figure}

We expand significantly fewer nodes by adopting a two-sided search approach. We hash the state vectors of a ball of radius $R$ around the goal state, and check every step for intersection. This is very fast, and we are able to pick $R$ up to 7 with no significant performance penalty.  Interestingly, $R$ needs to be sufficiently large to get good performance - for large beam width $k>12$, $R$ needs to be at least 3. We conjecture that this is related to the issue of $T$-calibration described in the text. See Fig. \ref{fig:scoresvsball} for more detail. We also experimented with a temperature parameter in the inverse process, but decreasing the temperature only hurt results.

One advantage--and perhaps disadvantage--of our method is that it considers the mathematically motivated diffusion time $T$. Accordingly, this becomes a hyperparameter that, for practical purposes, should be chosen at some value similar to the diameter of the group. For a sufficiently nice group, where the random walk mixing time is not too much larger than the diameter, this should be sufficient. This allows us to solve generic configurations. However, when we are trying to solve a particular with a shortest possible path, a na\"ive beam search from $t=T$ will prejudice towards walks that take the full time $T$ to reach the origin. The two-sided search significantly ameliorates this problem, as is demonstrated in fig \ref{fig:scoresvsball}, but we are also able to improve the result by running the search multiple times (which we call $T$-calibration). The argument is essentially that we should pick our start time $t$ as small as possible but not too small; so we start initially from $T$, and find a path of length $T'<T$ to the origin. We then run the inference again from $T'$, and iterate until no solution is found or $T'=T$. In practice, if the $T$ used in training is well-calibrated to the group, we usually only need to run twice, doubling execution time for a modest improvement in excess score and optimality. If we had perfect reconstruction of the score $\sigma^*$, such a strategy would yield the shortest path.

As discussed in \S \ref{s:graphs}, by twisting single cubelets
one can make ``invalid'' Rubik's cube positions which cannot be brought by cube moves to the
standard position, in other words live on different G-orbits.
How does the solver treat these?  We did this experiment by 
training on the standard G-orbit, and then solving from positions on the G-orbit
obtained by random walk from a configuration with a single twisted cubelet. Unfortunately, there is no good notion of a unique solved state when on a different orbit. Appropriate generalisation will have been achieved if the model is still able to bring the cube to a state which is near the goal state on an adjacent $G$ orbit, for some reasonable notion of adjacency. To judge performance we count all 8 configurations obtained by twisting each individual corner on a solved cube clockwise; in principle we should perhaps consider all combinations of $n$ twists with $n \mod 3 = 1$, but even this is a choice. Although these states are on the same orbit, whilst they are all `close' to the goal state, they are not necessarily close on their Cayley graph. In fact, evidence from various trained models suggests that they are all pairwise 14 moves apart, although we did not verify this with an optimal solver. We also tried the same problem but treating only the original twisted goal state as the new goal state. We use a ball of radius $R=5$ for each goal state.

By this definition of a solution, the average solution length at beam width $2^{18}$ was 20.81, significantly shorter than the equivalent on the usual orbit. This is presumably because there is a significant advantage in having these 8 different goal states. The dropoff in performance with beam width was rapid, however, as at beam width $2^{14}$ we fail to solve all states. If we instead define only one goal state, performance significantly degrades. Even at beam width $2^18$, we are now unable to solve 2.7\% of problems, with an average solution length of 25.94. This is certainly not surprising, given that there is nothing to distinguish this particular goal state.

\begin{figure}
    \centering
    \includegraphics[width=0.5\linewidth]{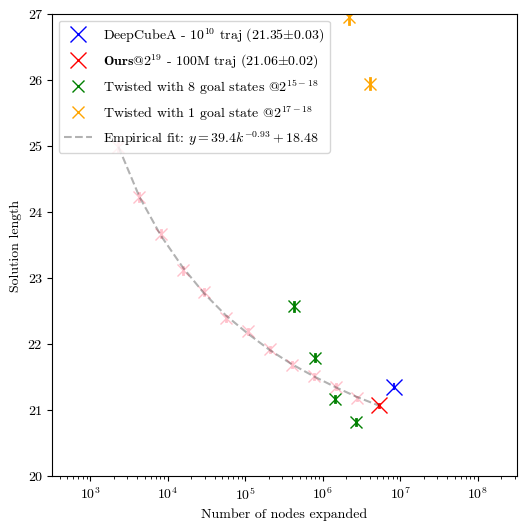}
    \caption{Even at beam width $18$, with 1 goal state we failed to solve all problems, so we present the mean just for illustration. With 8 goal states, we only failed to solve all states at beam width $2^{14}$ (not plotted).}
    \label{fig:twisted}
\end{figure}

\begin{figure}
    \centering
    \includegraphics[width=0.5\linewidth]{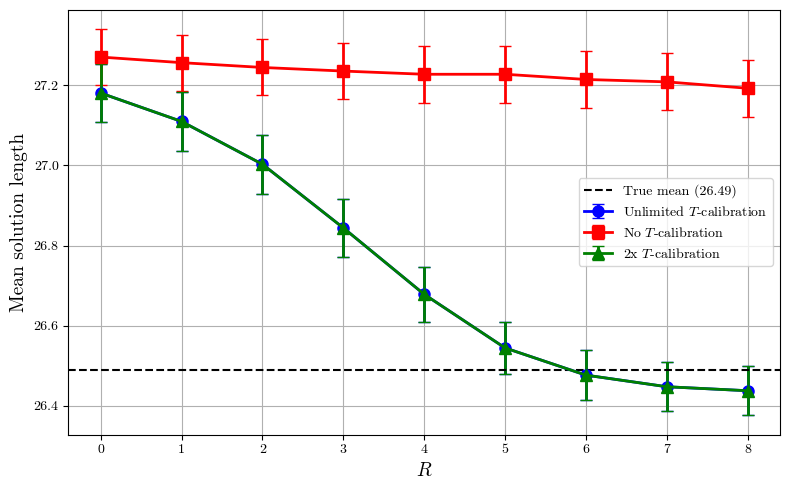}
    \caption{Dotted line is the true average distance. Plotted with the radius of the hashed target state ball, $R$. Varying effect of T-calibration, 1000 samples.}
        \label{fig:sl2}
\end{figure}

We also implemented a highly optimised breadth-first search algorithm to optimally solve $SL_2(\mathbb{Z}_p)$ for moderately large $p$.  We consider $p = 997$, cardinality 991,025,976. With our choice of generators ($T$ and $U$), the diameter is 39, and the average distance between states is 26.49. We also trained a neural network, similarly structured to the Rubik's cube, with 0.1M parameters on 1M examples and $T=50$. It was evaluated on 1000 states from the uniform distribution on $SL_2(\mathbb{Z}_{997})$. At the time of training the network, we did not know the diameter; it was therefore a good exercise in analysing the effect of $T$-calibration and hashed ball radius. See figure \ref{fig:sl2}. Unsurprisingly, when the ball size increases, the average solution length decreases. What is surprising is how much even one round of $T$-calibration affects performance. We don't see this effect nearly so much on the Rubik's cube because we have a much better calibrated $T$ to begin with. The average mean predicted by the network is consistent with the true value.

\section{Conclusions and questions for future work}
\label{s:conclusions}

We showed how to formulate graph navigation algorithms as diffusion models.
The exploration of the graph is carried out by a forward process, while finding the target nodes is done by
the inverse backward process.  This systematizes the discussion and suggests many generalizations.

We formulated and studied generalizations which vary the forward process to promote exploration.  The
``reversed score'' ansatz of \S \ref{ss:revscore} is particularly simple to implement.  In our experiments
it worked well, substantially improving over previous algorithms run with comparable resources.

While the diffusion model paradigm can be used for general undirected graphs, we focused 
on Cayley graphs of groups and group actions.  This is in part because of their many mathematical
applications, but also because these graphs have more structure on which to base generalization.
The mathematical understanding of groups is deep and it would be valuable to find ways to encode it
in the algorithms, perhaps by tailoring the score model $\sigma_\theta$.  

Many of the outstanding mathematical problems in this area reduce to determining whether a Cayley
graph of a group action is connected.  Another generalization which is natural in the diffusion model
paradigm (and used in almost all applications of it) is to start the forward process from multiple positions.  
In this case the reverse process will head towards the closer starting positions, or combinations 
of features from various positions.  This is vividly shown by image generation models, in which each
image in the training set is a starting position, and the generated images combine features from many
images (recent theoretical studies include \cite{kamb_analytic_2024,bradley_mechanisms_2025}).
Determining connectedness by looking for paths to the identity can work but is not very efficient; perhaps 
the use of multiple targets or other generalizations will make it more tractable.

\section*{Acknowledgments}

\indent

We thank Alexander Chervov, Jordan Ellenberg and Thomas Harvey for valuable discussions.  
This project was started at the Harvard CMSA Mathematics and Machine Learning Program held in fall 2024.
KFT would like to acknowledge the hospitality of IAIFI at the Massachusetts Institute of Technology, where a portion of this research was conducted. 

KFT is supported by the Gould-Watson Scholarship.

\appendix
\section{Shortest paths in $SL_2(\BZ)$ and $SL_2(\BZ_p)$}
\label{app:euc}


We start with a matrix $g=\left(\begin{matrix}a & b \\ c & d \end{matrix}\right)$.
If $b=0$, then $g=U_b = (U_1)^b$.  If $a=0$, then $g=ST_{-b}$.
If $a\ne 0$ and $b\ne 0$, then we take an iterative step which is guaranteed to decrease $\max(a,b)$.
Suppose $a\ge b$, then divide $a$ by $b$ with remainder, so $a=k_1 b+a_1$.  Then
\be
\left(\begin{matrix}a_1 & b \\ c - d k_1 & d \end{matrix}\right)
=
\left(\begin{matrix}a & b \\ c & d \end{matrix}\right)
\left(\begin{matrix}1 & 0 \\ -k_1 & 1 \end{matrix}\right)
\ee
is an application of $U_{-k_1}$.
The case $b>a$ is similar and leads to $T_{-k_1}$.

The same algorithm can be used for $SL_2(\BZ_p)$ by interpreting the matrix elements as integers
(``lifting the matrix to $SL_2(\BZ)$'') but doing this in a naive way does not produce the shortest path. 
In \cite{larsen2003navigating} this is adapted to a (probabilistic) n algorithm to find
paths of length $\CO(\log p \log\log p)$.  This does not quite match the diameter
$\CO(\log p)$ but is apparently the best known constructive result.


\bibliographystyle{alpha}
\bibliography{bibliography}

\address{
CMSA, Harvard University \\
20 Garden St, Cambridge MA 02138 \\
\email{mdouglas@cmsa.fas.harvard.edu}\\
}

\address{
Rudolf Peierls Centre for Theoretical Physics,
University of Oxford\\
Parks Road, Oxford OX1 3PU, UK \\
\email{cristofero.fraser-taliente@physics.ox.ac.uk}\\
}

\end{document}